\documentclass[sigconf]{acmart}

\AtBeginDocument{%
  \providecommand\BibTeX{{%
    \normalfont B\kern-0.5em{\scshape i\kern-0.25em b}\kern-0.8em\TeX}}}

\setcopyright{acmcopyright}
\copyrightyear{2018}
\acmYear{2018}
\acmDOI{XXXXXXX.XXXXXXX}

\acmConference[Conference acronym 'XX]{Make sure to enter the correct
  conference title from your rights confirmation emai}{June 03--05,
  2018}{Woodstock, NY}
\acmPrice{15.00}
\acmISBN{978-1-4503-XXXX-X/18/06}

\usepackage{booktabs}
\usepackage{enumitem}
\usepackage{xcolor}
\usepackage{subfigure}
\usepackage{multicol}
\usepackage{multirow}

\usepackage{algpseudocode}
\usepackage{amsfonts}
 
\usepackage{amssymb}
\usepackage{algorithm}

\usepackage{makecell}
\usepackage{xspace}
\usepackage{balance}

\newcommand{\modelname}{Fints\xspace}

\newcommand{\minisection}[1]{\vspace{0pt}\noindent\textbf{#1.}}

\begin{document}

\title{\modelname: Efficient Inference-Time Personalization for LLMs with Fine-Grained Instance-Tailored Steering}


\author{Kounianhua Du$^1$, Jianxing Liu$^1$, Kangning Zhang$^1$, 
 Wenxiang Jiao$^2$, Yuan Lu$^2$, \\
 Jiarui Jin$^2$, Weiwen Liu$^1$, Yong Yu$^1$, Weinan Zhang$^1$}
\affiliation{%
  \institution{$^1$Shanghai Jiao Tong University, $^2$ Xiaohongshu Inc.}
  \city{Shanghai}
  \country{China}
  }
\email{{kounianhuadu,	flweb3ranni,zhangkangning,wwliu,yyu,wnzhang}@sjtu.edu.cn}
\email{{wenxiangjiaonju,luyuan3,jinjiarui}@xiaohongshu.com}

\renewcommand{\shortauthors}{Kounianhua Du et al.}

\begin{abstract}
The rapid evolution of large language models (LLMs) has intensified the demand for effective personalization techniques that can adapt model behavior to individual user preferences. Despite the non-parametric methods utilizing the in-context learning ability of LLMs, recent parametric adaptation methods, including personalized parameter-efficient fine-tuning and reward modeling emerge. However, these methods face limitations in handling dynamic user patterns and high data sparsity scenarios, due to low adaptability and data efficiency. To address these challenges, we propose a fine-grained and instance-tailored steering framework that dynamically generates sample-level interference vectors from user data and injects them into the model's forward pass for personalized adaptation. 
 Our approach introduces two key technical innovations: a fine-grained steering component that captures nuanced signals by hooking activations from attention and MLP layers, and an input-aware aggregation module that synthesizes these signals into contextually relevant enhancements. The method demonstrates high flexibility and data efficiency, excelling in fast-changing distribution and high data sparsity scenarios. In addition, the proposed method is orthogonal to existing methods and operates as a plug-in component compatible with different personalization techniques. Extensive experiments across diverse scenarios—including short-to-long text generation, and web function calling—validate the effectiveness and compatibility of our approach. Results show that our method significantly enhances personalization performance in fast-shifting environments while maintaining robustness across varying interaction modes and context lengths. Implementation is available at https://github.com/KounianhuaDu/Fints.
\end{abstract}




\received{20 February 2007}
\received[revised]{12 March 2009}
\received[accepted]{5 June 2009}

\maketitle

\section{Introduction}
\label{sec:intro}
The advent of large language models (LLMs)~\cite{dubey2024llama,yang2025qwen3} has catalyzed transformative advances across a multitude of domains, from natural language understanding to complex reasoning tasks. A particularly compelling frontier within this landscape is the pursuit of personalization—the ability to tailor model behavior and outputs to individual user preferences~\cite{liu2025survey,cai2025large,wang2025never,salemi2023lamp,kumar2024longlamp}. 
\begin{figure}
    \centering
    \includegraphics[width=0.90\linewidth]{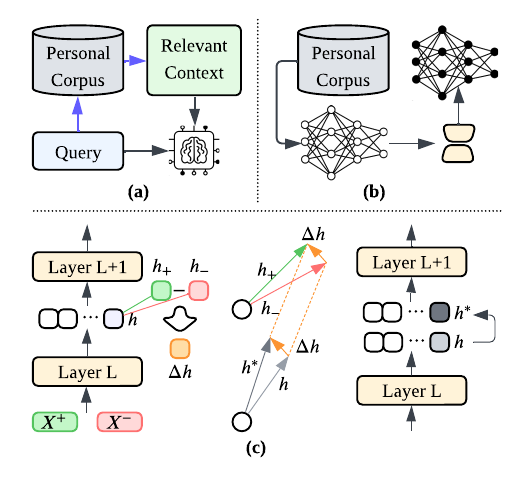}
    \caption{Illustration of different methodologies. (a) Prompt-based methods retrieve relevant context and feed it with the target query into the large language models (LLMs) for personalized output. (b) Personalized parameter-efficient tuning methods adapt LLMs with user data and obtain personalized weights to offer personalization. (c) Steering-based methodology constructs contrastive prompts from user logs and obtains a personal interference vector to guide model behavior.}
    \label{fig:intro}
\end{figure}
Achieving effective personalization is paramount for deploying LLMs in real-world applications such as personal assistant~\cite{wang2025roles,guha2015user}, conversational agents~\cite{yusuf2025pedagogical,liu2025proactive}, and adaptive learning systems~\cite{iqbal2025personalized,saleem2025ai}. 

Existing methodologies for personalization technology of LLM can be broadly categorized into three paradigms. Prompt-based methods leverage the in-context learning~\cite{dong2022survey} capabilities of LLMs by strategically inserting user profiles or historical interactions into the context window, guiding the model in a non-parametric manner~\cite{madaan2022memory,CFRAG,qian2024memorag}. Parametric adaption methods include One-Peft-All-User~\cite{lin2024rella} that learns collaborative signals within the whole dataset by tuning LLMs on all the users' data and One-Peft-Per-User~\cite{salemi2024optimization,tan2024personalized,PLoRA} that performs parameter-efficient fine-tuning on user-specific data, learning dedicated adapters that inject personalization directly into the model's weights. A third paradigm employs personalized reward models that act as judges, scoring multiple candidate generations to select the output most aligned with a user's preferences~\cite{zhuang2024hydra,chen2025pal,SynthesizeMe}. 

However, production environments expose two notoriously challenging regimes that severely limit the effectiveness of existing approaches:
\begin{itemize}[leftmargin=*]
\item \textbf{Extreme Data Sparsity}: New users often arrive with fewer than ten historical interactions, creating a cold-start~\cite{zhang2025cold} problem where traditional parametric methods fail due to insufficient training data.
\item \textbf{Fast-Changing User Patterns}: User preferences can shift dramatically within days or even hours~\cite{liu2024retrieval}, requiring personalization systems to adapt rapidly without expensive retraining cycles.
\end{itemize}
While prompt-based methods are flexible, they are fundamentally constrained by the finite context length of transformers~\cite{amazon2025context}, creating a bottleneck on the amount of personal data that can be utilized.
Parametric methods (personalized LoRAs), face critical challenges in dynamic environments:
1) Data Inefficiency: Require substantial user-specific data for training, performing poorly in low-data regimes. 2) Structural Interference: A single low-rank adapter is shared across all user prompts, causing gradient updates for new styles to overwrite previously encoded patterns—a \emph{local-minima forgetting} effect exacerbated by the low-rank bottleneck $(r=8-16)$. 3) Slow Adaptation: Bound by offline training cycles, making them ill-suited for rapidly evolving user preferences. Futhermore, researches~\cite{liu2025curse, yang2025resistance} show LLMs can possess an inherent "elasticity" or resistance to alignment changes after pre-training, implying that superficial parametric fine-tuning might struggle to effect lasting behavioral change.
Reward-based methods, while effective in stationary environments, they suffer from: 1) Action-Space Explosion. For example, in function-calling scenarios, the action space grows combinatorially with API arity; generating a sufficient number of candidates (e.g., $k!\geq!64$) per query to cover the space becomes computationally prohibitive.
2) Non-Stationary Bias. These methods assume a stationary utility function. In practice, the true reward surface can drift within a session (e.g., a user "adds to cart" then "removes"), producing biases that cannot be easily averaged out by mini-batch preference learning. This mirrors broader findings that LLMs can be sensitive to distribution shifts and may not generalize robustly beyond their training data patterns~\cite{jiang2024importance}.

\smallskip
In short, \emph{low-rank parameter sharing} and \emph{stationary reward priors}—the very design choices that make LoRA and reward models attractive in \emph{static} regimes—become their Achilles heel under fast, heterogeneous, and low-shot personal dynamics. 

\smallskip
To overcome the limitations, we introduce \textbf{\modelname}, an \textbf{inference-time steering} framework that treats personalization as a \textbf{sample-level activation shift rather than a parametric update}.
Steering vectors, by virtue of being \emph{training-free}, \emph{instance-selected}, and \emph{rank-unconstrained}, bypass all the pathologies above. The methodologies shift is illustrated in Figure~\ref{fig:intro}. 

Concretely, \modelname constructs a lightweight \emph{interference vector} per user query by aggregating contextually relevant steering vectors, which are prebuilt from contrasting model's internal activations under the positive (personal) prompts and the negative (impersonal) prompts.
Two technical designs empower \modelname in the aforementioned harsh regimes:
\textbf{fine-grained hooking} that separately extracts attention and MLP signals, capturing subtle style indicators that whole-layer steering overlooks, and \textbf{input-aware aggregation} that re-weights the most relevant historical steering vectors at run-time, delivering \emph{instant} adaptation.
Because the base model remains frozen and only a minimal vector is temporarily inserted to a mid-layer activation, \modelname incurs \emph{zero} gradient storage, \emph{zero} per-user checkpoints and light latency overhead. Emprical studies show that \modelname outperforms all the competing baselines across different validation scenarios and show extreme superiority in dynamic and data-sparse environments.

\smallskip
\noindent\textbf{Contributions} are summarized below:
\begin{itemize}[leftmargin=*]
  \item We propose \modelname, an \textbf{inference-time personalization} framework that delivers robust personalization by treating it as a sample-level activation shift. Particularly, \modelname shows distinct superiority under high data sparsity (\emph{$<$10-shot}) and fast interest drift, while bringing low memory overhead and adding light latency. 

  \item We propose a \textbf{fine-grained hooking} methodology that separately extracts attention and MLP activations, yielding $>$0.5\% Rouge-1 gains over whole-layer steering on both short- and long-text benchmarks.

  \item We propose an \textbf{input-aware aggregation} module that dynamically re-weights historical signals, enabling instance-level adaptation without re-training or re-deployment, offering flexibility and robustness.

  \item Systematic evaluation on scenarios including short content generation, long content generation, and web function calling. Various  ablation studies to validate \modelname's flexibility, distributional robustness, superior data efficiency, and light overhead. 
\end{itemize}

\section{Related Work and Preliminary}
\subsection{Prompt Based Methods}
Prompt-based methods primarily leverage the in-context learning capabilities of LLMs to achieve personalization in a non-parametric manner. These approaches involve retrieving relevant context from a user's personal content and profile to guide an LLM toward generating desired output:
\begin{equation}
    y=M_\theta \left(q||\phi(D_u)\right),
\end{equation}
where $\phi$ is a function that extracts relevant context from the user's personal context $D_u$, $||$ denotes the concatenation operation that fuses query $q$ and the relevant personalized context $\phi(D_u)$.

Based on different designs of $\phi$, methods can vary. Retrieval-augmented prompting excels at extracting the most relevant records from user data to enhance LLMs, utilizing user histories for personalization~\cite{salemi2023lamp}. Advanced methods further process this information by summarizing dense user profiles from interaction histories, offering compressed yet informative clues about user preference~\cite{cai2025large}. While highly flexible and requiring no model training, these methods are constrained by the finite context window of LLMs~\cite{vuong2021does}, which can bottleneck the amount of personal context that can be effectively utilized.

\subsection{Parametric Adaption Methods}
A distinct parametric approach involves light fine-tuning of a small subset of the LLM's parameters on a user's data. This is most commonly achieved through methods like Low-Rank Adaptation~\cite{hu2022lora}, which injects trainable rank-decomposition matrices into the model architecture.
\begin{equation}
    y = M_{\theta + \delta_u}\left(q\right),
\end{equation}
where $M_{\theta + \delta_u}$ denotes the model tailored for user $u$, $\theta$ is the shared parameter of the original LLM, and $\delta_u$ denotes the incremental personalized weights.

The methodologies of these methods can be categorized into two-folds: one-peft-all-users and one-peft-per-user. One-peft-all-user, e.g., Rella~\cite{lin2024rella}, finetunes large language models with all the user logs, capturing the collaborative signals within the dataset and adapting language models to follow specific instruction answering modes.
One-peft-per-user finetunes user-specific weights to achieve parametric personalization, achieving user-level adaption. 
OPPU~\cite{salemi2024optimization} integrates parametric user knowledge in the personal PEFT parameters with non-parametric knowledge from retrieval and profiles, adapting LLMs to user behavior shifts.
PER-PCS~\cite{tan2024personalized} turns personal LoRA parameters into a communal Lego set, where sharers contribute gated pieces and target users snap together the pieces that best match their history, obtaining a bespoke model without ever training a full personal LoRA or exposing raw private data.
Although more parameter-efficient than full fine-tuning, these methods still require a non-trivial amount of user-specific data for training and involve a persistent storage and loading overhead for each user's adapter weights. More critically, they are inherently slow to adapt, making them ill-suited for environments where user interests evolve rapidly.

\subsection{Personalized Reward Models}
Another line of work approaches personalization by training personalized reward models to evaluate and rank multiple candidate generations from an LLM. The core idea is to learn a user-specific scoring function that reflects individual preferences, which is then used to select the most suitable output from a set of candidates, often through techniques like rejection sampling~\cite{liu2023statistical} or best-of-$N$ sampling~\cite{gui2024bonbon}.

The technical realization of this paradigm typically involves training a lightweight scoring head or adapter for each user on their historical preference data. A prominent example is Hydra~\cite{zhuang2024hydra}, which trains a user-specific scoring adapter using pairs of positive and negative samples. A recent trend within this paradigm seeks to leverage the LLM itself as the personalized judge. By using carefully crafted prompts, the same base model can perform both generation and personalized evaluation, reducing the need for separate reward model training~\cite{wang2025never}.


While effective, these approaches often require generating multiple candidates per query, leading to significant computational overhead during inference, and rely on the availability of sufficient preference data for training robust reward models. Furthermore, accurately rewarding model generations under complex, real-world scenarios is intrinsically difficult due to the vast action space, necessitating vast and diverse-enough data for proper coverage~\cite{synthesizeme2025stanford}.

\subsection{Formulation of Steering-Based Personalization}
Our work is distinct in its core mechanism. Unlike prompt-based methods, we operate through direct activation steering rather than context space. Unlike reward models, we avoid the computational cost of multiple generations and subsequent ranking. Unlike fine-tuning methods, our inference-time vectors require no gradient-based training and offer instant adaptability, positioning our approach as a highly flexible and data-efficient alternative for dynamic personalization. Additionally, the methodologies are orthogonal to the competing baselines, serving as a compatible plug-in to different personalization techniques and offering instant adaption where other methods cannot.

Concretely, steering intervenes in a \emph{frozen} LLM $\mathcal{M}_\theta$ by adding a learned vector $\Delta$ to an intermediate activation $h_l$:
\begin{equation}
\tilde{h}_l = h_l + \gamma \cdot\Delta,
\quad \text{where } h_l=M_\theta^l(X),\; \gamma\in\mathbb{R}.
\end{equation}
The vector $\Delta\in\mathbb{R}^d$ is \emph{fixed} for the user (or sample) and is \emph{not} updated via gradients; only the activation is modified during the forward pass, keeping $\theta$ unchanged. 

This formulation rests on the linear representation hypothesis~\cite{mikolov2013efficient}, which posits that latent features are captured as linear directions in the representation space; consequently, adjusting activations by adding linear shift vectors can steer the model's predictions toward a target behavior.
This \emph{linear shift} is sufficient because Transformer blocks are piece-wise linear: the offset propagates through FFNs and attention, directly biasing key-query similarities and value mixtures 
Empirically, even a single-layer constant perturbation reliably moves output distributions along desired attributes (style, sentiment, persona) without cascading instability.  
Thus, ``\emph{linearly shift latent representations}'' offers a training-free, computationally trivial, yet effective mechanism for inference-time LLM control~\cite{turner2023steering,frandsen2022extracting}. 

Steering-based personalization therefore reduces to (i) selecting the injection layer $l$ and scale $\gamma$; (ii) constructing an appropriate $\Delta$ from user data.
For the formal problem, empirical studies recommend scanning a \emph{single} intermediate-to-late layer (typically $l\!\approx\!0.4\!-\!0.6L$) on a small validation set and choosing the earliest layer that maximizes the target metric while keeping generation perplexity unchanged~\cite{turner2023steering,li2025enhancing,anonymous2025chain,li2025enhancing}. 
With the chosen $l$ fixed, $\gamma$ is grid-searched in $[0.05,0.8]$ (step $0.05$); the value that yields the best validation score without significant log-probability drop is adopted.  
Once $(l,\gamma)$ are determined, they are frozen for test-time deployment; distribution drift can be handled by re-tuning only $\gamma$ on new validation data, leaving $l$ unchanged.  
This minimal search procedure keeps the entire personalization pipeline training-free and rapid.
The latter problem, which is the focus of steering based personalization, will be detailed in the methodology part.

\begin{figure*}[!ht]
    \centering
    \includegraphics[width=0.95\linewidth]{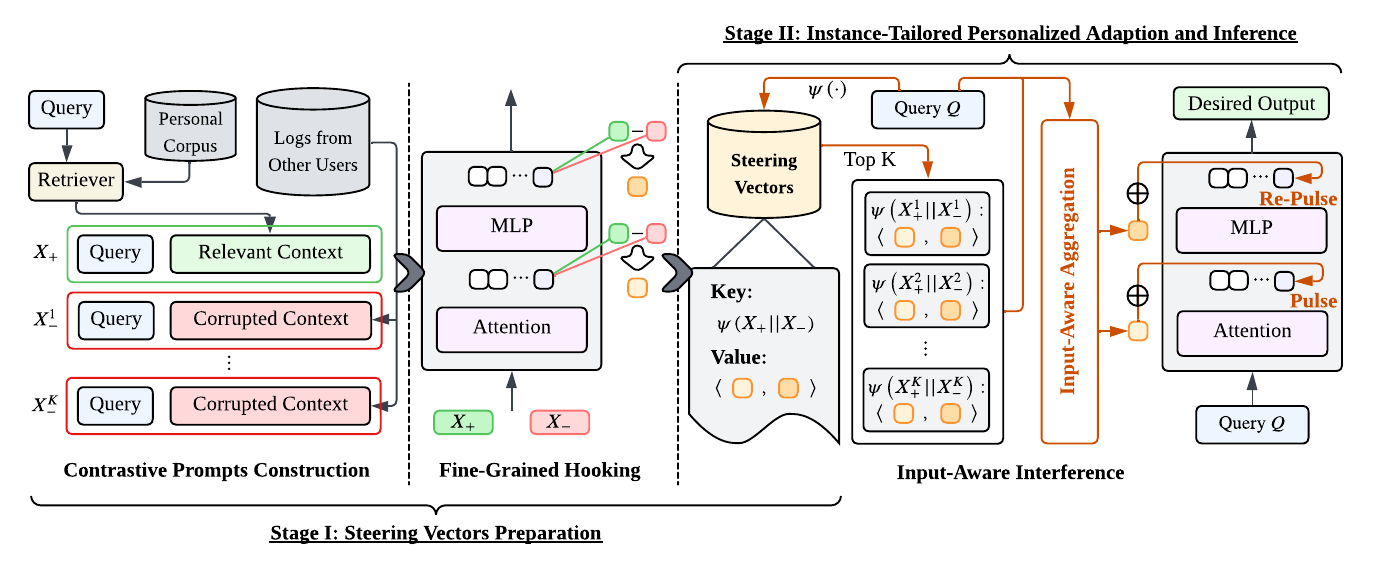}
    \vspace{-15pt}
    \caption{Overview of \modelname. 1) Steering Vectors Preparation. During this stage, we construct contrastive prompts from user logs, where relevant context retrieved from personal corpus is concatenated with target query to form the positive sample and irrelevant context sampled from other users is concatenated with target query to form the negative sample. For each positive sample, we generate $K$ negative samples. Each pair is then fed into the LLM in the teacher forcing mode, during which the last token representation of attention and MLP blocks are hooked. We store the difference between two sample activations of a pair to serve as the steering vector, with the text of each pair being the key for indexing convenience. 2) Instance-Tailored Personalized Adaption and Inference. During this stage, we sample from target user's steering vectors set to interfere model for personalized output. Concretely, we rank the similarity between query text and sample pair text to select the top-k steering vectors, which are then attentively aggregated and injected into LLM for personalized adaption.}
    \label{fig:framework}
\end{figure*}

\section{Methodology}
\subsection{Overview}



Despite the increasing demand of personalized LLMs service, existing methods face critical problems in dynamic and data-sparse environments. Limitations of existing methods include structural interference, slow adaptation, and context window constraints, which hinder their applicability in real-world scenarios with fast-changing user interests and limited interaction data.

In this paper, we propose \modelname, a novel inference-time personalization framework that treats personalization as a sample-level activation shift rather than a parametric update. \modelname operates through a two-stage pipeline designed for instance-aware personalization without model training:

\begin{itemize}[leftmargin=*]
    \item \textbf{Offline Steering Vectors Preparation}: For each user, we construct a personalized dictionary of steering vectors from historical interactions. Each vector is derived by contrasting model activations under user-relevant and user-irrelevant contexts, using a \textit{fine-grained hooking} mechanism that separately extracts signals from attention and MLP layers.
    
    \item \textbf{Online Inference with Input-Aware Steering}: During inference, for a target query, Fints dynamically selects and aggregates the most relevant steering vectors from the user's dictionary based on semantic similarity. The aggregated vectors are then injected into the model's forward pass at a predefined layer via a \textit{Pulse and Re-Pulse} mechanism, enabling real-time, query-specific personalization. This approach ensures that each query receives a tailored activation shift, offering instance-level adaption. The two-step injection---\textit{Pulse} (after attention) and \textit{Re-Pulse} (after MLP)--- captures subtle signals from each component and offers reinforced interference.
\end{itemize}

These designs enable \modelname to adapt instantly with fine-grained signals to meet evolving user interests, with minimal memory and latency overhead, making it particularly suitable for dynamic and data-sparse environments where traditional parametric methods struggle. 

\subsection{Steering Vectors Preparation}
To obtain parametric interference vectors for each user, we construct steering vectors from historical user logs.
\subsubsection{Contrastive Prompts Construction}
For a historical sample $X$, we construct a set of contrastive prompts $\{\langle X_+, X_-\rangle\}$ to obtain the steering vectors, where $X_+$ contains relevant context resulting to desired output and $X_-$ is corrupted with irrelevant context from other users resulting to undesired output:
\begin{align}
    C_+ = &Retriever(Q, D_u),\\
    X_+ = &(Q||C_+),\\
    C_- = &RandomSample(D_{\setminus u}),\\
    X_- = &(Q||C_-),
\end{align}
where $D_u$ denotes the set of historical question-answer pairs of user $u$, $D_{\setminus u}$ denotes the set of historical question-answer pairs from users excluding $u$, $C_+$ denotes the relevant context for $X$, and $C_-$ denotes the corrupted context for $X$.

\subsubsection{Fine-Grained Hooking}
For each pair of sample $\{\langle X_+, X_-\rangle\}$, we then compute $f(X_+)$ and $f(X_-)$ in teacher forcing mode, during which we extract the latent representation at a layer $L$ for the last generated tokens. To capture the subtle signals, we hook activations from the attention block and MLP layer separately:
\begin{align}
    h^{L_{attn}}_+ = M^L_{\theta_{attn}}(X_+),\quad &h^{L_{mlp}}_+ = M^L_{\theta_{mlp}}\left(M^L_{\theta_{attn}}(X_+)\right),\\
    h^{L_{attn}}_- = M^L_{\theta_{attn}}(X_-),\quad  &h^{L_{mlp}}_- = M^L_{\theta_{mlp}}\left(M^L_{\theta_{attn}}(X_-)\right),
\end{align}

where $M^L_{\theta_{attn}}$ denotes the attention module and $M^L_{\theta_{mlp}}$ denotes the mlp module.

We then obtain the difference between the two representations to obtain the steering vector
\begin{align}
    \Delta_{attn} = &h^{L_{attn}}_+ - h^{L_{attn}}_-,\\
    \Delta_{mlp} = &h^{L_{mlp}}_+ - h^{L_{mlp}}_-,
\end{align}

For each historical sample of user $u$, we generate $K$ pairs of contrastive prompts $\{\langle X_+^{u_i}, X_-^{u_i}\rangle|_{i=1}^K\}$ by combining different corrupted contexts with query to serve as the negative samples and obtain the corresponding steering vectors as above.
For indexing convenience, we use the encoding of the sample pair as the key and use their corresponding steering vector as the value:
\begin{align}
    \forall i\in\{1,\dots,K\}, \quad &\mathbf{k}^{u_i} = \psi(X^{u_i}_+|| X^{u_i}_-)\label{eq:encoding}, \\
    \mathcal{S}_u = \{\mathbf{k}^{u_i}&: \langle\Delta_{attn}^{u_i}, \Delta_{mlp}^{u_i}\rangle\},
\end{align}
where $\psi(\cdot)$ is an sequence encoding model and $\mathcal{S}_u$ denotes the steering vectors dict for user $u$.

\subsection{Instance-Tailored Personalized Adaption and Inference}

To achieve flexible and context-aware personalization, \modelname employs an input-aware interference mechanism during inference. This approach dynamically selects and aggregates steering vectors based on the semantic similarity between the target query and historical user interactions, enabling instance-level adaptation without requiring model retraining or parameter updates.

\subsubsection{Input-Aware Aggregation}

The aggregation process begins by computing the similarity between the target query and each entry in the user's steering vector dictionary. For each key $\mathbf{k}^{u_i} \in \mathcal{S}_u$, we calculate the cosine distance:
\begin{equation}
d_i = 1 - \text{cosine}(\psi(Q), \mathbf{k}^{u_i}), \quad \forall i \in \{1,\ldots,|\mathcal{S}_u|\},
\end{equation}
where $\psi(\cdot)$ denotes the same sequence encoding model employed during steering vector preparation (equation \ref{eq:encoding}), ensuring consistent representation spaces for similarity computation.

We then identify the top-K most relevant sample pairs by selecting those with minimal distances:
\begin{equation}
\{\mathbf{k}^{uj}, \langle\Delta^{uj}_{attn}, \Delta^{uj}_{mlp}\rangle\}_{j=1}^{K} \xleftarrow{\text{Top K min}} \{d_i\}_{i=1}^{|\mathcal{S}_u|}.
\end{equation}

The selected steering vectors are aggregated through one of the following methods:
\begin{itemize}
    \item \textbf{Mean Aggregation}:
    \begin{equation}
    \mathbf{s}_{attn} = \frac{1}{K}\sum_{j=1}^{K}\Delta_{attn}^{uj}, \quad \mathbf{s}_{mlp} = \frac{1}{K}\sum_{j=1}^{K}\Delta_{mlp}^{uj}.
    \end{equation}
    
    \item \textbf{Attentive Aggregation}:
    \begin{align}
    w_j = &\frac{1-d_j}{\sum_{k=1}^{K}(1-d_k)}, \\
    \mathbf{s}_{attn} = \sum_{j=1}^{K}w_j \cdot \Delta_{attn}^{uj}&, \quad \mathbf{s}_{mlp} = \sum_{j=1}^{K}w_j \cdot \Delta_{mlp}^{uj}.
     \end{align}
\end{itemize}

The attentive aggregation scheme assigns higher weights to steering vectors derived from historical samples that are more semantically similar to the current query, enabling more precise personalization.

\subsubsection{Inference-Time Application: Pulse and Re-Pulse}
During the forward pass for target query $Q$ at the predefined injection layer $L$, the aggregated steering vectors are applied sequentially to respect the Transformer architecture:

\begin{itemize}
    \item Compute attention output at layer $L$:
    \begin{equation}
        h^{L}_{attn} = M^{L}_{\theta_{attn}}(Q).
    \end{equation}
    \item Pulse: Apply attention steering vector at layer $L$:
    \begin{equation}
        h^{L'}_{attn} = h^{L}_{attn} + \alpha \cdot s_{attn}.
    \end{equation}
    \item Compute MLP output at layer $L$:
    \begin{equation}
        h^{L}_{mlp} = M^{L}_{\theta_{mlp}}(h^{L'}_{attn}).
    \end{equation}
    \item Re-Pulse: Apply MLP steering vector at layer $L$:
    \begin{equation}
        h^{L'}_{mlp} = h^{L}_{mlp} + \beta \cdot s_{mlp}.
    \end{equation}
\end{itemize}

The final adjusted representation $h^{L'}_{mlp}$ propagates through subsequent layers $L+1$ to $L_{total}$ to generate the personalized output:
\begin{equation}
y = M^{L+1:L_{total}}_{\theta}(h^{L'}_{mlp}).
\end{equation}

The scaling factors $\alpha$ and $\beta$ control the steering intensity for attention and MLP components respectively, providing fine-grained control over the personalization effect while maintaining generation quality.

\begin{table*}[!t]
\centering
\caption{Main results on content generation and personal web function calling.}
\vspace{-10pt}
\label{tab:main}
\begin{tabular}{cccc|cc|cc|c}
\hline
\multicolumn{4}{c|}{Datasets}                                                                                                                                         & \multicolumn{2}{c|}{Headline Generation} & \multicolumn{2}{c|}{Abstract Writing} & PersonalWAB \\ \hline
\multicolumn{4}{c|}{Methods}                                                                                                                                          & Rouge-1        & \multicolumn{1}{c|}{Rouge-L}       & Rouge-1                 & Rouge-L             & ACC   \\ \hline
\multicolumn{1}{c|}{Direct}                               & \multicolumn{3}{c|}{ZeroShot}                                                                           & 0.1432         & \multicolumn{1}{c|}{0.1297}        & 0.3503                  & 0.2024                & 0.6017\\ \hline
\multicolumn{1}{c|}{\multirow{4}{*}{\makecell[c]{In Context\\ Learning}}} & \multicolumn{3}{c|}{k=1}                                                                                & 0.1389         & \multicolumn{1}{c|}{0.1287}        & 0.3648                  & 0.2093               &0.6767  \\
\multicolumn{1}{c|}{}                                       & \multicolumn{3}{c|}{k=3}                                                                                & 0.1643         & \multicolumn{1}{c|}{0.1515}        & 0.3701                  & 0.2104            &\underline{0.8415}     \\
\multicolumn{1}{c|}{}                                       & \multicolumn{3}{c|}{k=5}                                                                                & 0.1665         & \multicolumn{1}{c|}{0.1506}        & 0.3771                  & 0.2153               &0.8330  \\
\multicolumn{1}{c|}{}                                   & \multicolumn{3}{c|}{k=10}          &       0.1630         & 0.1488              &         0.3768                &     0.2127              &0.7966     \\ \hline
\multicolumn{1}{c|}{\multirow{3}{*}{\makecell[c]{Parametric Adaption}}}  & \multicolumn{3}{c|}{Rella}  &0.1635&	0.1463	&		0.3818&	0.2141&		0.8330 \\
\multicolumn{1}{c|}{}   & \multicolumn{3}{c|}{OPPU}                                                                               & \underline{0.1779}         & \multicolumn{1}{c|}{\underline{0.1613}}        &         \underline{0.3848}	&\underline{0.2146}            &0.8394    \\
\multicolumn{1}{c|}{}                                       & \multicolumn{3}{c|}{PER-PCS}                                                                            & 0.1763         & \multicolumn{1}{c|}{0.1577}        &         0.3602                &  0.2169               &0.8402       \\ \hline
\multicolumn{1}{c|}{\multirow{6}{*}{\modelname}}            & \multicolumn{3}{c|}{Naive Aggregation}                                            & 0.1737                   & 0.1575                   &      0.3946                   &          0.2277         &0.8458     \\
\multicolumn{1}{c|}{}                                       & \multicolumn{3}{c|}{Input Awareness}                    &     0.1768                     &       0.1601                   &         0.3982                &       0.2274          &0.8544       \\\cline{2-9} 
\multicolumn{1}{c|}{}                                       & \multicolumn{2}{c|}{\multirow{4}{*}{\makecell[c]{Fine-Grained \\Hooking}}} & attn      & 0.1752                   & 0.159                    & 0.3938                  & 0.2271           &\textbf{0.8588}      \\
\multicolumn{1}{c|}{}                                       & \multicolumn{2}{c|}{}                                     & mlp                   & 0.1704                   & 0.1544                   & 0.3826                  & 0.2133           &0.8501      \\ 
\multicolumn{1}{c|}{}                                       & \multicolumn{2}{c|}{}                                     & whole                 & 0.1737                   & 0.1575                   & 0.3946                  & 0.2277             &0.8458    \\ \cline{4-9}
\multicolumn{1}{c|}{}                                       & \multicolumn{2}{c|}{}                                     & attn + mlp            & \textbf{0.1816}                   & \textbf{0.1666}                   &       \textbf{0.3990}                  &        \textbf{0.2306}              &0.8522  \\ \hline
\end{tabular}
\end{table*}

\section{Experiment}
\subsection{Setup}
\subsubsection{Datasets}
To evaluate \modelname under different scenarios, we utilize datasets including short personalized content generation (headline generation\footnote{https://lamp-benchmark.github.io/}), long personalized content generation (abstract writing\footnote{https://lamp-benchmark.github.io/}), and personal web function calling (PersonalWAB\footnote{https://github.com/HongruCai/PersonalWAB/tree/main/PersonalWAB}) for evaluation. Each of them contains user id, corresponding profile, and interaction histories. 

\begin{table}[h]
\centering
\caption{Statistics of Used Datasets.}
\vspace{-10pt}
\label{tab:sta}
\begin{tabular}{c|cccc}
\hline

Datasets                                                                            & \#Users & \makecell[c]{Avg. \\Profiles} & \makecell[c]{Avg. Text\\Length \\(per user)} & \makecell[c]{Avg. Text\\Length \\(per sample)} \\ \hline
\makecell[c]{Headline\\Generation}                                                         & 200     & 68.26         & 2257                                                                  & 33                                                                      \\\hline
\makecell[c]{Abstract\\Writing}                                                            & 200     & 117.91        & 18369                                                                 & 156                                                                     \\\hline
\begin{tabular}[c]{@{}c@{}}PersonalWAB\\ (product)\end{tabular}     & 200     & 60.72         & 20348                                                                 & 335                                                                     \\\hline
\begin{tabular}[c]{@{}c@{}}PersonalWAB\\ (interaction)\end{tabular} & 200     & 12.2          & 656                                                                   & 54                                                                      \\ \hline
\end{tabular}
\end{table}
\begin{itemize}[leftmargin=10pt]
    \item \textbf{News-Headline Generation}. This dataset~\cite{misra2022news} evaluates the ability of a large language model to capture the stylistic patterns of an author, querying the LLM to generate a headline given the input news article. 
    \item \textbf{Abstract Writing}. This dataset evaluates LLMs' ability to distill complex ideas and generate accurate, concise, and coherent output over the span of multiple paragraphs on domain-specific tasks. The input of the model is the title of the paper along with some keywords to guide the content. The expected output is an abstract conditioned on the title and keywords in the user's style\cite{kumar2024longlamp}.
    \item  \textbf{PersonalWAB}. This dataset integrates personalized instruction comprehension and action execution, where LLMs
    must infer personalized user requirements and preferences to determine which Web function to call and formulate the corresponding function parameters. Subsequently, the results of these function calls are returned to users\cite{cai2025large}.
\end{itemize}

Based on the time-split setting, we select the top-200 users with the most sufficient interaction histories, where earlier question-answer~(QA) pairs form the train set and the latter pairs form the test set. The statistics of used datasets are summarized in Table~\ref{tab:sta}.

\subsubsection{Baselines}
We evaluate \modelname against: 1) prompt-based method (In Context Learning) and 2) parametric adaption methods including One-Peft-All-User (Rella\cite{lin2024rella}) and One-Peft-Per-User (OPPU\cite{salemi2024optimization}, PER-PCS\cite{tan2024personalized}). Personalized reward models are out-of-scope since they involve generating multiple answers to pick from, where extra tokens counts poses unfair comparison. 

\subsubsection{Metric}
We use Rouge Values \cite{lin2004rouge} to measure the quality of content generation and use Overall Accuracy (ACC) to measure the quality of web function calling.

\subsection{Research Questions}
We evaluate methods starting from the research questions below.
\begin{itemize}[leftmargin=10pt]
    \item \textbf{(RQ1)} Does \modelname outperform the competing baselines?
    \item \textbf{(RQ2)} Is each design component of \modelname effective?
    \item  \textbf{(RQ3)} How is the robustness of \modelname against heterogeneous/fast changing distribution compared to other methods?
    \item \textbf{(RQ4)} How is the data efficiency of \modelname compared to other methods? Does \modelname still perform well under high data sparsity?
    \item \textbf{(RQ5)} How is the overhead brought by \modelname? 
\end{itemize}

\subsection{Main Results (RQ1-RQ2)}
The main results of the experiments are included in Table~\ref{tab:main}.

\minisection{\textbf{(RQ1)}} From the results, we can see that \modelname outperforms all other baselines. Concretely, parametric methods show better performances than the prompt-based methods, showcasing the benefits of personalization in parameters to better digest user patterns. Futhermore, \modelname outperforms traditional personalized loras methods, justifying the superiority of the steering-based methodology.

\begin{table*}[!t]
\caption{Results of different methods against heterogeneous distribution.}
\vspace{-10pt}
\label{tab:he}
\begin{tabular}{cc|cc|cc|cc}
\hline
\multicolumn{2}{c|}{Datasets}                                        & \multicolumn{2}{c|}{Sub-Population 1} & \multicolumn{2}{c|}{Sub-Population 2} & \multicolumn{2}{c}{Sub-Population 3} \\ \hline
\multicolumn{2}{c|}{Methods}                                         & Rouge-1           & Rouge-L           & Rouge-1           & Rouge-L           & Rouge-1           & Rouge-L          \\ \hline
\multicolumn{2}{c|}{In Context Learning}                                             & 0.1598            & 0.1469            & 0.1965            & 0.1808            & 0.1574            & 0.1427           \\ \hline
\multicolumn{1}{c|}{\multirow{3}{*}{Parametric Adaption}} & Rella    & 0.1520            & 0.1379            & 0.1607            & 0.1422            & 0.1582            & 0.1393           \\
\multicolumn{1}{c|}{}                                     & OPPU     & 0.1605            & 0.1452            & 0.1766            & 0.1663            & 0.1749            & 0.1566           \\
\multicolumn{1}{c|}{}                                     & PER-PCS   & 0.1434            & 0.1249            & 0.1644            & 0.1588            & 0.1546            & 0.1428           \\ \hline
\multicolumn{1}{c|}{\multirow{4}{*}{\modelname}}            & attn   & 0.168             & 0.1511            & 0.1931            & 0.1755            & 0.1677            & 0.1544           \\
\multicolumn{1}{c|}{}                                     & mlp    & 0.1627            & 0.1454            & 0.196             & 0.1748            & 0.1608            & 0.1445           \\ 
\multicolumn{1}{c|}{}                                     & whole  & 0.1762            & 0.1567            & 0.2047            & 0.1868            & 0.1766            & 0.1642           \\\cline{2-8} 
\multicolumn{1}{c|}{}                                     & attn+mlp & 0.1700            & 0.1523            & 0.1847            & 0.1648            & 0.1639            & 0.1413           \\ \hline
\end{tabular}
\end{table*}

\minisection{\textbf{(RQ2)}} The last five lines of the table display the performances of variants of \modelname, from which we can draw the following conclusions: 1) input-aware aggregation performs better than the direct mean aggregation, showcasing significant performance gains over the three datasets; 2) overall, fine-grained hooking captures more subtle signals than whole module hooking or separate component hooking, bringing significant performance gain ($>0.5\%$) in the contents generation datasets. These observations justify the effectiveness of the components of \modelname.

\subsection{Case Study}
In this section, we conduct various case studies to justify our method.

\subsubsection{(RQ3) Robustness to Heterogeneous Distribution} 
\begin{figure}[!h]
    \centering
    \includegraphics[width=0.80\linewidth]{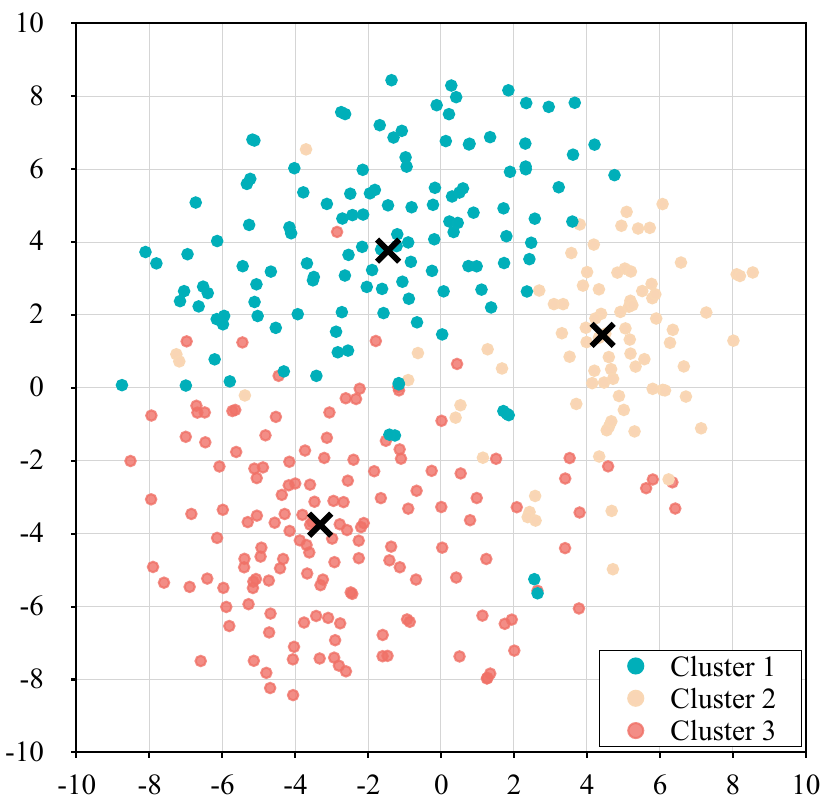}
    \vspace{-10pt}
    \caption{T-SNE visualization of the data distribution, where the heterogeneous test sets are sampled from.}
    \label{fig:tsne}
\end{figure}

In this section, we study the robustness of parametric methods over heterogeneous distributions. Concretely, we first find sub-populations existing in the data, and then sample data points from each sub-population, over which different methods are tested on. The T-SNE visualization~\cite{maaten2008visualizing} of different sub-populations is illustrated in Figure~\ref{fig:tsne}, where datapoints of different sub-populations are visualized in different colors. We sample points near each cluster center to form the heterogeneous test sets.

\begin{figure}
    \centering
    \includegraphics[width=0.95\linewidth]{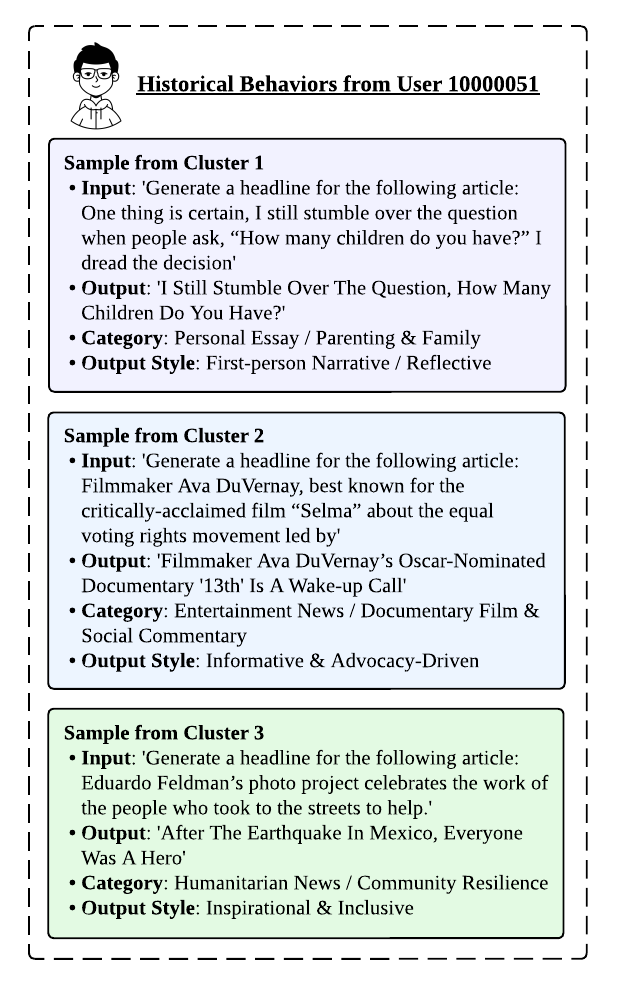}
    \vspace{-15pt}
    \caption{Illustration of the heterogeneous data.}
    \label{fig:case}
\end{figure}

To further illustrate the discrepancy existing in the heterogeneous test sets, we select one sample from each cluster of the same user $10000051$ as illustrated in Figure~\ref{fig:case}.
These illustrations demonstrate the heterogeneous content themes, audience engagement strategies, and emotional tones existing in Question-Answer pairs of the same user. We then conduct evaluation on the heterogeneous test sets to test the robustness of different methods. The results of different methods are summarized in Table~\ref{tab:he}.

\begin{table*}[!t]
\caption{Data efficiency analysis of different methods.}
\vspace{-10pt}
\label{tab:df}
\begin{tabular}{cc|cc|cc|cc|cc}
\hline
\multicolumn{2}{c|}{\#Used Data}                                    & \multicolumn{2}{c|}{5\% (Avg. 5.82)}                       & \multicolumn{2}{c|}{15\% (Avg. 17.46)}                     & \multicolumn{2}{c|}{25\% (Avg. 29.10)}                     & \multicolumn{2}{c}{50\% (Avg. 58.72)}                     \\ \hline
\multicolumn{2}{c|}{Methods}                                        & \multicolumn{1}{c}{Rouge-1} & \multicolumn{1}{c|}{Rouge-L} & \multicolumn{1}{c}{Rouge-1} & \multicolumn{1}{c|}{Rouge-L} & \multicolumn{1}{c}{Rouge-1} & \multicolumn{1}{c|}{Rouge-L} & \multicolumn{1}{c}{Rouge-1} & \multicolumn{1}{c}{Rouge-L} \\ \hline
\multirow{2}{*}{Personalized Loras} & OPPU                          & 0.3764                      & 0.2000                       & 0.3788                      & 0.2033                       & 0.3794                      & 0.2132                       & 0.3834  & 0.2200  \\
                                    & PER-PCS                        & 0.3600                      & 0.2154                       & 0.3644                      & 0.2158                       & 0.3597                      & 0.2150                       & 0.3594                      & 0.2152                      \\ \hline
\multirow{4}{*}{\modelname}           & attn                          & 0.3925                      & 0.2234                       & 0.3912                      & 0.2229                       & 0.3921                      & 0.2266                       & 0.3950                      & 0.2277                      \\
                                    & whole                         & 0.3928                      & 0.2240                       & 0.3945                      & 0.2256                       & 0.3962                      & 0.2283                       & 0.3967                      & 0.2282                      \\
                                    & mlp                           & 0.3828                      & 0.2144                       & 0.3811                      & 0.2118                       & 0.3832                      & 0.2135                       & 0.3836                      & 0.2132                      \\ \cline{2-10} 
                                    & \multicolumn{1}{l|}{attn+mlp} & 0.3919                      & 0.2236                       & 0.3895                      & 0.2211                       & 0.3956                      & 0.2254                       & 0.3973                      & 0.2268                      \\ \hline
\end{tabular}
\end{table*}

From the results, we can see that \modelname outperforms the personalized lora methods in each sub-population. Since the personalized loras cannot quickly realign to rapidly changing user preferences due to their lock to sluggish and offline training loops. While \modelname offers inference-time  and instance-tailored adaption, and therefore it offers flexibility and robustness to heterogeneous distributions. This feature makes \modelname distinctive in scenario where user patterns change fast, which is a commonly-seen and important problem in LLMs personalization.

\subsubsection{(RQ4) Data Efficiency}
\begin{figure}[h]
    \centering
    \includegraphics[width=0.9\linewidth]{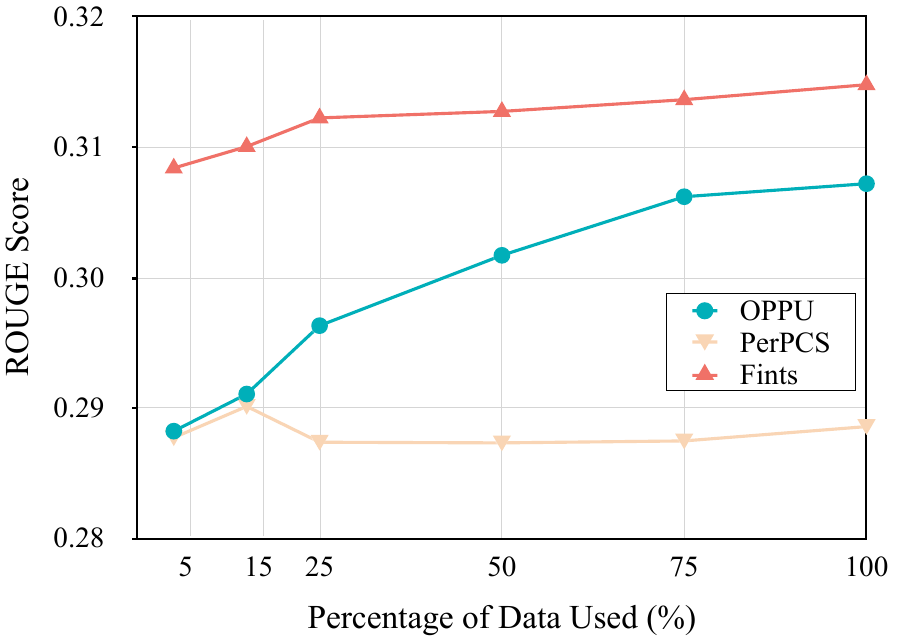}
    \vspace{-10pt}
    \caption{Data efficiency analysis of different methods.}
    \label{fig:df}
    \vspace{-10pt}
\end{figure}

Despite the robustness of heterogeneous distribution, data sparsity is also an important problem in LLMs personalization, where only a small amount of data is available to conduct effective personalization. Therefore, we study the data efficiency of different methods in this section. Concretely, we use different amount of personalized data to train models/serve as steering materials. The results are summarized in Table~\ref{tab:df} and visualized in Figure~\ref{fig:df}.

Generally, lora-based methods either maintain a consistently weak performance or encounter a drastic performance drop as we keep reducing the amount of data. In contrast, our \modelname is insensitive to data decrease. Particularly, under an extreme scenario where less than 10 personalized user logs are provided, \modelname still maintains its superior performance and outperforms lora-based methods by up to 3\% absolute Rouge score. 
This observation distinctly validates the superiority of \modelname under conditions of high data sparsity, where other methods struggle.

\subsubsection{(RQ5) Overhead Discussion}
To further study the overhead brought by \modelname, we sample 200 datapoints from each dataset and conduct the evaluation. The results are displayed in Table~\ref{tab:time}.

\begin{table}[h]
\centering
\caption{Latency Analysis.}
\vspace{-10pt}
\label{tab:time}
\begin{tabular}{c|cc}
\toprule
\multirow{2}{*}{Datasets (200 samples)} & \multicolumn{2}{c}{Headline Generation}        \\ 
                                        & Direct Inference & + \modelname \\ \hline
Latency (s)                             & 63.23            & 72.49                       \\ \hline
&&  \\\\[-20pt]\hline
\multirow{2}{*}{Datasets (200 samples)} & \multicolumn{2}{c}{Abstract Writing}           \\ 
                                        & Direct Inference & + \modelname \\ \hline
Latency (s)                             & 670.34           & 712.36                      \\ \hline
&&  \\\\[-20pt] \hline 
\multirow{2}{*}{Datasets (200 samples)} & \multicolumn{2}{c}{PersonalWAB}                \\ 
                                        & Direct Inference & + \modelname \\ \hline
Latency (s)                             & 1161.88          & 1343.89                     \\ \bottomrule
\end{tabular}
\end{table}

The inference overhead for \modelname is primarily attributed to retrieving the relevant steering vectors from the set, which is the most time-consuming step, and subsequently inserting the aggregated steering signal into the LLM's forward pass. While latency varies with the steering vector size per user and the query's token count. Table~\ref{tab:time} displays the results of time latency analysis brought by \modelname, which is conducted on a single NVIDIA H100 device. Results demonstrate that \modelname incurs only a light overhead during inference.  This efficiency offers a crucial advantage by circumventing the huge upload burden typically associated with transferring distinct personalized weights. 


\section{Conclusion}
In this paper, we present \modelname, a novel inference-time personalization framework that addresses the critical challenges of data efficiency and adaptability in LLM personalization. Unlike traditional parametric methods that require extensive fine-tuning, \modelname operates through activation steering at inference time, making it particularly suitable for scenarios with fast-changing user preferences and limited training data.
The core innovations of \modelname include: (1) \textit{fine-grained hooking} that separately extracts steering signals from attention and MLP layers, capturing nuanced user preferences that whole-layer approaches miss; and (2) \textit{input-aware aggregation} that dynamically weights historical steering vectors based on semantic similarity to the current query, enabling instance-level adaptation without retraining.
Extensive evaluation across three diverse tasks—news headline generation, scientific abstract writing, and personalized web function calling—demonstrates \modelname's superior performance over both prompt-based and parameter-efficient fine-tuning baselines. Notably, \modelname maintains strong performance even under extreme data scarcity (as few as 5-6 examples per user) and rapidly shifting user distributions, scenarios where traditional methods exhibit significant performance degradation.

Practical advantages of \modelname include minimal memory overhead, negligible latency increase, and seamless compatibility with existing personalization techniques as a plug-in component. These characteristics make \modelname particularly well-suited for production environments where storage, computation, and adaptability constraints are paramount.

\begin{acks}
To Robert, for the bagels and explaining CMYK and color spaces.
\end{acks}

\bibliographystyle{ACM-Reference-Format}
\bibliography{sample-base}


\begin{thebibliography}{43}


\ifx \showCODEN    \undefined \def \showCODEN     #1{\unskip}     \fi
\ifx \showDOI      \undefined \def \showDOI       #1{#1}\fi
\ifx \showISBNx    \undefined \def \showISBNx     #1{\unskip}     \fi
\ifx \showISBNxiii \undefined \def \showISBNxiii  #1{\unskip}     \fi
\ifx \showISSN     \undefined \def \showISSN      #1{\unskip}     \fi
\ifx \showLCCN     \undefined \def \showLCCN      #1{\unskip}     \fi
\ifx \shownote     \undefined \def \shownote      #1{#1}          \fi
\ifx \showarticletitle \undefined \def \showarticletitle #1{#1}   \fi
\ifx \showURL      \undefined \def \showURL       {\relax}        \fi
\providecommand\bibfield[2]{#2}
\providecommand\bibinfo[2]{#2}
\providecommand\natexlab[1]{#1}
\providecommand\showeprint[2][]{arXiv:#2}

\bibitem[Anonymous(2025)]%
        {anonymous2025chain}
\bibfield{author}{\bibinfo{person}{Anonymous}.} \bibinfo{year}{2025}\natexlab{}.
\newblock \showarticletitle{Activation Steering for Chain-of-Thought Compression}.
\newblock \bibinfo{journal}{\emph{OpenReview preprint}} (\bibinfo{year}{2025}).
\newblock


\bibitem[Cai et~al\mbox{.}(2025)]%
        {cai2025large}
\bibfield{author}{\bibinfo{person}{Hongru Cai}, \bibinfo{person}{Yongqi Li}, \bibinfo{person}{Wenjie Wang}, \bibinfo{person}{Fengbin Zhu}, \bibinfo{person}{Xiaoyu Shen}, \bibinfo{person}{Wenjie Li}, {and} \bibinfo{person}{Tat-Seng Chua}.} \bibinfo{year}{2025}\natexlab{}.
\newblock \showarticletitle{Large language models empowered personalized web agents}. In \bibinfo{booktitle}{\emph{Proceedings of the ACM on Web Conference 2025}}. \bibinfo{pages}{198--215}.
\newblock


\bibitem[Chen et~al\mbox{.}(2025)]%
        {chen2025pal}
\bibfield{author}{\bibinfo{person}{Daiwei Chen}, \bibinfo{person}{Yi Chen}, \bibinfo{person}{Aniket Rege}, \bibinfo{person}{Zhi Wang}, {and} \bibinfo{person}{Ramya~Korlakai Vinayak}.} \bibinfo{year}{2025}\natexlab{}.
\newblock \showarticletitle{{PAL}: Sample-Efficient Personalized Reward Modeling for Pluralistic Alignment}. In \bibinfo{booktitle}{\emph{The Thirteenth International Conference on Learning Representations}}.
\newblock
\urldef\tempurl%
\url{https://openreview.net/forum?id=1kFDrYCuSu}
\showURL{%
\tempurl}


\bibitem[Dong et~al\mbox{.}(2022)]%
        {dong2022survey}
\bibfield{author}{\bibinfo{person}{Qingxiu Dong}, \bibinfo{person}{Lei Li}, \bibinfo{person}{Damai Dai}, \bibinfo{person}{Ce Zheng}, \bibinfo{person}{Jingyuan Ma}, \bibinfo{person}{Rui Li}, \bibinfo{person}{Heming Xia}, \bibinfo{person}{Jingjing Xu}, \bibinfo{person}{Zhiyong Wu}, \bibinfo{person}{Tianyu Liu}, {et~al\mbox{.}}} \bibinfo{year}{2022}\natexlab{}.
\newblock \showarticletitle{A survey on in-context learning}.
\newblock \bibinfo{journal}{\emph{arXiv preprint arXiv:2301.00234}} (\bibinfo{year}{2022}).
\newblock


\bibitem[Dubey et~al\mbox{.}(2024)]%
        {dubey2024llama}
\bibfield{author}{\bibinfo{person}{Abhimanyu Dubey}, \bibinfo{person}{Abhinav Jauhri}, \bibinfo{person}{Abhinav Pandey}, \bibinfo{person}{Abhishek Kadian}, \bibinfo{person}{Ahmad Al-Dahle}, \bibinfo{person}{Aiesha Letman}, \bibinfo{person}{Akhil Mathur}, \bibinfo{person}{Alan Schelten}, \bibinfo{person}{Amy Yang}, \bibinfo{person}{Angela Fan}, {et~al\mbox{.}}} \bibinfo{year}{2024}\natexlab{}.
\newblock \showarticletitle{The llama 3 herd of models}.
\newblock \bibinfo{journal}{\emph{arXiv e-prints}} (\bibinfo{year}{2024}), \bibinfo{pages}{arXiv--2407}.
\newblock


\bibitem[Frandsen et~al\mbox{.}(2022)]%
        {frandsen2022extracting}
\bibfield{author}{\bibinfo{person}{Abraham Frandsen}, \bibinfo{person}{Rong Ge}, {and} \bibinfo{person}{Vatsal Sharan}.} \bibinfo{year}{2022}\natexlab{}.
\newblock \showarticletitle{Extracting Latent State Representations with Linear Dynamics from Language Models}. In \bibinfo{booktitle}{\emph{Proc. of ICML}}.
\newblock


\bibitem[Guha et~al\mbox{.}(2015)]%
        {guha2015user}
\bibfield{author}{\bibinfo{person}{Ramanathan Guha}, \bibinfo{person}{Vineet Gupta}, \bibinfo{person}{Vivek Raghunathan}, {and} \bibinfo{person}{Ramakrishnan Srikant}.} \bibinfo{year}{2015}\natexlab{}.
\newblock \showarticletitle{User modeling for a personal assistant}. In \bibinfo{booktitle}{\emph{Proceedings of the Eighth ACM International Conference on Web Search and Data Mining}}. \bibinfo{pages}{275--284}.
\newblock


\bibitem[Gui et~al\mbox{.}(2024)]%
        {gui2024bonbon}
\bibfield{author}{\bibinfo{person}{Lin Gui}, \bibinfo{person}{Cristina G{\^a}rbacea}, {and} \bibinfo{person}{Victor Veitch}.} \bibinfo{year}{2024}\natexlab{}.
\newblock \showarticletitle{Bonbon alignment for large language models and the sweetness of best-of-n sampling}.
\newblock \bibinfo{journal}{\emph{Advances in Neural Information Processing Systems}}  \bibinfo{volume}{37} (\bibinfo{year}{2024}), \bibinfo{pages}{2851--2885}.
\newblock


\bibitem[Hu et~al\mbox{.}(2022)]%
        {hu2022lora}
\bibfield{author}{\bibinfo{person}{Edward~J Hu}, \bibinfo{person}{Yelong Shen}, \bibinfo{person}{Phillip Wallis}, \bibinfo{person}{Zeyuan Allen-Zhu}, \bibinfo{person}{Yuanzhi Li}, \bibinfo{person}{Shean Wang}, \bibinfo{person}{Lu Wang}, \bibinfo{person}{Weizhu Chen}, {et~al\mbox{.}}} \bibinfo{year}{2022}\natexlab{}.
\newblock \showarticletitle{Lora: Low-rank adaptation of large language models.}
\newblock \bibinfo{journal}{\emph{ICLR}} \bibinfo{volume}{1}, \bibinfo{number}{2} (\bibinfo{year}{2022}), \bibinfo{pages}{3}.
\newblock


\bibitem[Iqbal et~al\mbox{.}(2025)]%
        {iqbal2025personalized}
\bibfield{author}{\bibinfo{person}{Muddesar Iqbal}, \bibinfo{person}{Sohail Sarwar}, \bibinfo{person}{Muhammad Safyan}, {and} \bibinfo{person}{Moustafa Nasralla}.} \bibinfo{year}{2025}\natexlab{}.
\newblock \showarticletitle{Personalized and adaptive e-learning systems for semantic Web: a systematic review and roadmap}.
\newblock \bibinfo{journal}{\emph{International Journal of Web Information Systems}} \bibinfo{volume}{21}, \bibinfo{number}{4} (\bibinfo{year}{2025}), \bibinfo{pages}{327--352}.
\newblock


\bibitem[Jiang et~al\mbox{.}(2024)]%
        {jiang2024importance}
\bibfield{author}{\bibinfo{person}{Yifan Jiang} {et~al\mbox{.}}} \bibinfo{year}{2024}\natexlab{}.
\newblock \showarticletitle{Importance Weighting Can Help Large Language Models Self-Improve}.
\newblock \bibinfo{journal}{\emph{arXiv preprint arXiv:2408.09849}} (\bibinfo{year}{2024}).
\newblock


\bibitem[Kumar et~al\mbox{.}(2024)]%
        {kumar2024longlamp}
\bibfield{author}{\bibinfo{person}{Ishita Kumar}, \bibinfo{person}{Snigdha Viswanathan}, \bibinfo{person}{Sushrita Yerra}, \bibinfo{person}{Alireza Salemi}, \bibinfo{person}{Ryan~A Rossi}, \bibinfo{person}{Franck Dernoncourt}, \bibinfo{person}{Hanieh Deilamsalehy}, \bibinfo{person}{Xiang Chen}, \bibinfo{person}{Ruiyi Zhang}, \bibinfo{person}{Shubham Agarwal}, {et~al\mbox{.}}} \bibinfo{year}{2024}\natexlab{}.
\newblock \showarticletitle{Longlamp: A benchmark for personalized long-form text generation}.
\newblock \bibinfo{journal}{\emph{arXiv preprint arXiv:2407.11016}} (\bibinfo{year}{2024}).
\newblock


\bibitem[Li and Others(2025)]%
        {li2025enhancing}
\bibfield{author}{\bibinfo{person}{Shizheng Li} {and} \bibinfo{person}{Others}.} \bibinfo{year}{2025}\natexlab{}.
\newblock \showarticletitle{Enhancing Instruction Following of Language Models via Activation Steering}. In \bibinfo{booktitle}{\emph{Proc. of ICLR}}.
\newblock


\bibitem[Lin(2004)]%
        {lin2004rouge}
\bibfield{author}{\bibinfo{person}{Chin-Yew Lin}.} \bibinfo{year}{2004}\natexlab{}.
\newblock \showarticletitle{Rouge: A package for automatic evaluation of summaries}. In \bibinfo{booktitle}{\emph{Text summarization branches out}}. \bibinfo{pages}{74--81}.
\newblock


\bibitem[Lin et~al\mbox{.}(2024)]%
        {lin2024rella}
\bibfield{author}{\bibinfo{person}{Jianghao Lin}, \bibinfo{person}{Rong Shan}, \bibinfo{person}{Chenxu Zhu}, \bibinfo{person}{Kounianhua Du}, \bibinfo{person}{Bo Chen}, \bibinfo{person}{Shigang Quan}, \bibinfo{person}{Ruiming Tang}, \bibinfo{person}{Yong Yu}, {and} \bibinfo{person}{Weinan Zhang}.} \bibinfo{year}{2024}\natexlab{}.
\newblock \showarticletitle{Rella: Retrieval-enhanced large language models for lifelong sequential behavior comprehension in recommendation}. In \bibinfo{booktitle}{\emph{Proceedings of the ACM Web Conference 2024}}. \bibinfo{pages}{3497--3508}.
\newblock


\bibitem[Liu et~al\mbox{.}(2025c)]%
        {liu2025survey}
\bibfield{author}{\bibinfo{person}{Jiahong Liu}, \bibinfo{person}{Zexuan Qiu}, \bibinfo{person}{Zhongyang Li}, \bibinfo{person}{Quanyu Dai}, \bibinfo{person}{Jieming Zhu}, \bibinfo{person}{Minda Hu}, \bibinfo{person}{Menglin Yang}, {and} \bibinfo{person}{Irwin King}.} \bibinfo{year}{2025}\natexlab{c}.
\newblock \showarticletitle{A survey of personalized large language models: Progress and future directions}.
\newblock \bibinfo{journal}{\emph{arXiv preprint arXiv:2502.11528}} (\bibinfo{year}{2025}).
\newblock


\bibitem[Liu et~al\mbox{.}(2024)]%
        {liu2024retrieval}
\bibfield{author}{\bibinfo{person}{Jianghao Liu}, \bibinfo{person}{Rong Shan}, \bibinfo{person}{Chenxu Zhu}, \bibinfo{person}{Kounanhua Du}, \bibinfo{person}{Bo Chen}, \bibinfo{person}{Shigang Quan}, \bibinfo{person}{Ruiming Tang}, \bibinfo{person}{Yong Yu}, {and} \bibinfo{person}{Weinan Zhang}.} \bibinfo{year}{2024}\natexlab{}.
\newblock \showarticletitle{Relia: Retrieval-enhanced Large Language Model for Lifelong Sequential Behavior Comprehension in Recommendation}. In \bibinfo{booktitle}{\emph{Proceedings of the ACM Web Conference 2024}}. \bibinfo{pages}{3497--3508}.
\newblock


\bibitem[Liu et~al\mbox{.}(2025a)]%
        {liu2025curse}
\bibfield{author}{\bibinfo{person}{Shiwei Liu} {et~al\mbox{.}}} \bibinfo{year}{2025}\natexlab{a}.
\newblock \showarticletitle{The Curse of Depth in Large Language Models}.
\newblock \bibinfo{journal}{\emph{arXiv preprint arXiv:2502.05795v2}} (\bibinfo{year}{2025}).
\newblock


\bibitem[Liu et~al\mbox{.}(2023)]%
        {liu2023statistical}
\bibfield{author}{\bibinfo{person}{Tianqi Liu}, \bibinfo{person}{Yao Zhao}, \bibinfo{person}{Rishabh Joshi}, \bibinfo{person}{Misha Khalman}, \bibinfo{person}{Mohammad Saleh}, \bibinfo{person}{Peter~J Liu}, {and} \bibinfo{person}{Jialu Liu}.} \bibinfo{year}{2023}\natexlab{}.
\newblock \showarticletitle{Statistical rejection sampling improves preference optimization}.
\newblock \bibinfo{journal}{\emph{arXiv preprint arXiv:2309.06657}} (\bibinfo{year}{2023}).
\newblock


\bibitem[Liu et~al\mbox{.}(2025b)]%
        {liu2025proactive}
\bibfield{author}{\bibinfo{person}{Xingyu~Bruce Liu}, \bibinfo{person}{Shitao Fang}, \bibinfo{person}{Weiyan Shi}, \bibinfo{person}{Chien-Sheng Wu}, \bibinfo{person}{Takeo Igarashi}, {and} \bibinfo{person}{Xiang'Anthony' Chen}.} \bibinfo{year}{2025}\natexlab{b}.
\newblock \showarticletitle{Proactive conversational agents with inner thoughts}. In \bibinfo{booktitle}{\emph{Proceedings of the 2025 CHI Conference on Human Factors in Computing Systems}}. \bibinfo{pages}{1--19}.
\newblock


\bibitem[Maaten and Hinton(2008)]%
        {maaten2008visualizing}
\bibfield{author}{\bibinfo{person}{Laurens van~der Maaten} {and} \bibinfo{person}{Geoffrey Hinton}.} \bibinfo{year}{2008}\natexlab{}.
\newblock \showarticletitle{Visualizing data using t-SNE}.
\newblock \bibinfo{journal}{\emph{Journal of machine learning research}} \bibinfo{volume}{9}, \bibinfo{number}{Nov} (\bibinfo{year}{2008}), \bibinfo{pages}{2579--2605}.
\newblock


\bibitem[Madaan et~al\mbox{.}(2022)]%
        {madaan2022memory}
\bibfield{author}{\bibinfo{person}{Aman Madaan}, \bibinfo{person}{Niket Tandon}, \bibinfo{person}{Peter Clark}, {and} \bibinfo{person}{Yiming Yang}.} \bibinfo{year}{2022}\natexlab{}.
\newblock \showarticletitle{Memory-assisted prompt editing to improve GPT-3 after deployment}.
\newblock \bibinfo{journal}{\emph{arXiv preprint arXiv:2201.06009}} (\bibinfo{year}{2022}).
\newblock


\bibitem[Mikolov et~al\mbox{.}(2013)]%
        {mikolov2013efficient}
\bibfield{author}{\bibinfo{person}{Tomas Mikolov}, \bibinfo{person}{Kai Chen}, \bibinfo{person}{Greg Corrado}, {and} \bibinfo{person}{Jeffrey Dean}.} \bibinfo{year}{2013}\natexlab{}.
\newblock \showarticletitle{Efficient estimation of word representations in vector space}.
\newblock \bibinfo{journal}{\emph{arXiv preprint arXiv:1301.3781}} (\bibinfo{year}{2013}).
\newblock


\bibitem[Misra(2022)]%
        {misra2022news}
\bibfield{author}{\bibinfo{person}{Rishabh Misra}.} \bibinfo{year}{2022}\natexlab{}.
\newblock \showarticletitle{News category dataset}.
\newblock \bibinfo{journal}{\emph{arXiv preprint arXiv:2209.11429}} (\bibinfo{year}{2022}).
\newblock


\bibitem[Qian et~al\mbox{.}(2024)]%
        {qian2024memorag}
\bibfield{author}{\bibinfo{person}{Hongjin Qian}, \bibinfo{person}{Peitian Zhang}, \bibinfo{person}{Zheng Liu}, \bibinfo{person}{Kelong Mao}, {and} \bibinfo{person}{Zhicheng Dou}.} \bibinfo{year}{2024}\natexlab{}.
\newblock \showarticletitle{Memorag: Moving towards next-gen rag via memory-inspired knowledge discovery}.
\newblock \bibinfo{journal}{\emph{arXiv preprint arXiv:2409.05591}}  \bibinfo{volume}{1} (\bibinfo{year}{2024}).
\newblock


\bibitem[Ryan et~al\mbox{.}(2025a)]%
        {synthesizeme2025stanford}
\bibfield{author}{\bibinfo{person}{Michael~J. Ryan} {et~al\mbox{.}}} \bibinfo{year}{2025}\natexlab{a}.
\newblock \showarticletitle{SynthesizeMe: Understanding Users via Personalized Reward Models}.
\newblock \bibinfo{journal}{\emph{arXiv preprint}} (\bibinfo{year}{2025}).
\newblock
\newblock
\shownote{Available at arXiv:2506.05598v1}.


\bibitem[Ryan et~al\mbox{.}(2025b)]%
        {SynthesizeMe}
\bibfield{author}{\bibinfo{person}{Michael~J Ryan}, \bibinfo{person}{Omar Shaikh}, \bibinfo{person}{Aditri Bhagirath}, \bibinfo{person}{Daniel Frees}, \bibinfo{person}{William Held}, {and} \bibinfo{person}{Diyi Yang}.} \bibinfo{year}{2025}\natexlab{b}.
\newblock \bibinfo{title}{SynthesizeMe! Inducing Persona-Guided Prompts for Personalized Reward Models in LLMs}.
\newblock
\newblock
\showeprint[arxiv]{2506.05598}~[cs.CL]
\urldef\tempurl%
\url{https://arxiv.org/abs/2506.05598}
\showURL{%
\tempurl}


\bibitem[Saleem et~al\mbox{.}(2025)]%
        {saleem2025ai}
\bibfield{author}{\bibinfo{person}{Sadaf Saleem}, \bibinfo{person}{Muhammad~Umar Aziz}, \bibinfo{person}{Muhammad~Jawed Iqbal}, {and} \bibinfo{person}{Shahid Abbas}.} \bibinfo{year}{2025}\natexlab{}.
\newblock \showarticletitle{AI in education: Personalized learning systems and their impact on student performance and engagement}.
\newblock \bibinfo{journal}{\emph{The Critical Review of Social Sciences Studies}} \bibinfo{volume}{3}, \bibinfo{number}{1} (\bibinfo{year}{2025}), \bibinfo{pages}{2445--2459}.
\newblock


\bibitem[Salemi et~al\mbox{.}(2024)]%
        {salemi2024optimization}
\bibfield{author}{\bibinfo{person}{Alireza Salemi}, \bibinfo{person}{Surya Kallumadi}, {and} \bibinfo{person}{Hamed Zamani}.} \bibinfo{year}{2024}\natexlab{}.
\newblock \showarticletitle{Optimization methods for personalizing large language models through retrieval augmentation}. In \bibinfo{booktitle}{\emph{Proceedings of the 47th International ACM SIGIR Conference on Research and Development in Information Retrieval}}. \bibinfo{pages}{752--762}.
\newblock


\bibitem[Salemi et~al\mbox{.}(2023)]%
        {salemi2023lamp}
\bibfield{author}{\bibinfo{person}{Alireza Salemi}, \bibinfo{person}{Sheshera Mysore}, \bibinfo{person}{Michael Bendersky}, {and} \bibinfo{person}{Hamed Zamani}.} \bibinfo{year}{2023}\natexlab{}.
\newblock \showarticletitle{Lamp: When large language models meet personalization}.
\newblock \bibinfo{journal}{\emph{arXiv preprint arXiv:2304.11406}} (\bibinfo{year}{2023}).
\newblock


\bibitem[Science and Teams(2025)]%
        {amazon2025context}
\bibfield{author}{\bibinfo{person}{Amazon Science} {and} \bibinfo{person}{Research Teams}.} \bibinfo{year}{2025}\natexlab{}.
\newblock \showarticletitle{Context Length Alone Hurts LLM Performance Despite Perfect Retrieval}.
\newblock \bibinfo{journal}{\emph{Amazon Science Publications}} (\bibinfo{year}{2025}).
\newblock
\urldef\tempurl%
\url{https://www.amazon.science/publications/context-length-alone-hurts-llm-performance-despite-perfect-retrieval}
\showURL{%
\tempurl}


\bibitem[Shi et~al\mbox{.}(2025)]%
        {CFRAG}
\bibfield{author}{\bibinfo{person}{Teng Shi}, \bibinfo{person}{Jun Xu}, \bibinfo{person}{Xiao Zhang}, \bibinfo{person}{Xiaoxue Zang}, \bibinfo{person}{Kai Zheng}, \bibinfo{person}{Yang Song}, {and} \bibinfo{person}{Han Li}.} \bibinfo{year}{2025}\natexlab{}.
\newblock \showarticletitle{Retrieval Augmented Generation with Collaborative Filtering for Personalized Text Generation}. In \bibinfo{booktitle}{\emph{Proceedings of the 48th International ACM SIGIR Conference on Research and Development in Information Retrieval}} (Padua, Italy) \emph{(\bibinfo{series}{SIGIR '25})}. \bibinfo{publisher}{Association for Computing Machinery}, \bibinfo{address}{New York, NY, USA}, \bibinfo{pages}{1294–1304}.
\newblock
\showISBNx{9798400715921}
\urldef\tempurl%
\url{https://doi.org/10.1145/3726302.3730075}
\showDOI{\tempurl}


\bibitem[Tan et~al\mbox{.}(2024)]%
        {tan2024personalized}
\bibfield{author}{\bibinfo{person}{Zhaoxuan Tan}, \bibinfo{person}{Zheyuan Liu}, {and} \bibinfo{person}{Meng Jiang}.} \bibinfo{year}{2024}\natexlab{}.
\newblock \showarticletitle{Personalized pieces: Efficient personalized large language models through collaborative efforts}.
\newblock \bibinfo{journal}{\emph{arXiv preprint arXiv:2406.10471}} (\bibinfo{year}{2024}).
\newblock


\bibitem[Turner et~al\mbox{.}(2023)]%
        {turner2023steering}
\bibfield{author}{\bibinfo{person}{Alexander~M Turner}, \bibinfo{person}{Laura Thiergart}, \bibinfo{person}{Gabriel Leech}, \bibinfo{person}{David Udell}, \bibinfo{person}{Juan~J Vazquez}, \bibinfo{person}{Ulisse Mini}, {and} \bibinfo{person}{Monte MacDiarmid}.} \bibinfo{year}{2023}\natexlab{}.
\newblock \showarticletitle{Steering Language Models with Activation Engineering}.
\newblock \bibinfo{journal}{\emph{arXiv preprint arXiv:2308.10248}} (\bibinfo{year}{2023}).
\newblock


\bibitem[Vuong et~al\mbox{.}(2021)]%
        {vuong2021does}
\bibfield{author}{\bibinfo{person}{Tung Vuong}, \bibinfo{person}{Salvatore Andolina}, \bibinfo{person}{Giulio Jacucci}, {and} \bibinfo{person}{Tuukka Ruotsalo}.} \bibinfo{year}{2021}\natexlab{}.
\newblock \showarticletitle{Does more context help? Effects of context window and application source on retrieval performance}.
\newblock \bibinfo{journal}{\emph{ACM Transactions on Information Systems (TOIS)}} \bibinfo{volume}{40}, \bibinfo{number}{2} (\bibinfo{year}{2021}), \bibinfo{pages}{1--40}.
\newblock


\bibitem[Wang et~al\mbox{.}(2025)]%
        {wang2025never}
\bibfield{author}{\bibinfo{person}{Haoming Wang}, \bibinfo{person}{Boyuan Yang}, \bibinfo{person}{Xiangyu Yin}, {and} \bibinfo{person}{Wei Gao}.} \bibinfo{year}{2025}\natexlab{}.
\newblock \showarticletitle{Never Start from Scratch: Expediting On-Device LLM Personalization via Explainable Model Selection}.
\newblock \bibinfo{journal}{\emph{arXiv preprint arXiv:2504.13938}} (\bibinfo{year}{2025}).
\newblock


\bibitem[Wang and Sun(2025)]%
        {wang2025roles}
\bibfield{author}{\bibinfo{person}{Shaofeng Wang} {and} \bibinfo{person}{Zhuo Sun}.} \bibinfo{year}{2025}\natexlab{}.
\newblock \showarticletitle{Roles of artificial intelligence experience, information redundancy, and familiarity in shaping active learning: Insights from intelligent personal assistants}.
\newblock \bibinfo{journal}{\emph{Education and Information Technologies}} \bibinfo{volume}{30}, \bibinfo{number}{2} (\bibinfo{year}{2025}), \bibinfo{pages}{2525--2546}.
\newblock


\bibitem[Yang et~al\mbox{.}(2025b)]%
        {yang2025qwen3}
\bibfield{author}{\bibinfo{person}{An Yang}, \bibinfo{person}{Anfeng Li}, \bibinfo{person}{Baosong Yang}, \bibinfo{person}{Beichen Zhang}, \bibinfo{person}{Binyuan Hui}, \bibinfo{person}{Bo Zheng}, \bibinfo{person}{Bowen Yu}, \bibinfo{person}{Chang Gao}, \bibinfo{person}{Chengen Huang}, \bibinfo{person}{Chenxu Lv}, {et~al\mbox{.}}} \bibinfo{year}{2025}\natexlab{b}.
\newblock \showarticletitle{Qwen3 technical report}.
\newblock \bibinfo{journal}{\emph{arXiv preprint arXiv:2505.09388}} (\bibinfo{year}{2025}).
\newblock


\bibitem[Yang et~al\mbox{.}(2025a)]%
        {yang2025resistance}
\bibfield{author}{\bibinfo{person}{Yaodong Yang} {et~al\mbox{.}}} \bibinfo{year}{2025}\natexlab{a}.
\newblock \showarticletitle{Language Models Resist Alignment: Evidence From Data Compression}. In \bibinfo{booktitle}{\emph{Proceedings of the 63rd Annual Meeting of the Association for Computational Linguistics (ACL 2025)}}.
\newblock


\bibitem[Yusuf et~al\mbox{.}(2025)]%
        {yusuf2025pedagogical}
\bibfield{author}{\bibinfo{person}{Habeeb Yusuf}, \bibinfo{person}{Arthur Money}, {and} \bibinfo{person}{Damon Daylamani-Zad}.} \bibinfo{year}{2025}\natexlab{}.
\newblock \showarticletitle{Pedagogical AI conversational agents in higher education: a conceptual framework and survey of the state of the art}.
\newblock \bibinfo{journal}{\emph{Educational technology research and development}} \bibinfo{volume}{73}, \bibinfo{number}{2} (\bibinfo{year}{2025}), \bibinfo{pages}{815--874}.
\newblock


\bibitem[Zhang et~al\mbox{.}(2025)]%
        {zhang2025cold}
\bibfield{author}{\bibinfo{person}{Weizhi Zhang}, \bibinfo{person}{Yuanchen Bei}, \bibinfo{person}{Liangwei Yang}, \bibinfo{person}{Henry~Peng Zou}, \bibinfo{person}{Peilin Zhou}, \bibinfo{person}{Aiwei Liu}, \bibinfo{person}{Yinghui Li}, \bibinfo{person}{Hao Chen}, \bibinfo{person}{Jianling Wang}, \bibinfo{person}{Yu Wang}, {et~al\mbox{.}}} \bibinfo{year}{2025}\natexlab{}.
\newblock \showarticletitle{Cold-start recommendation towards the era of large language models (llms): A comprehensive survey and roadmap}.
\newblock \bibinfo{journal}{\emph{arXiv preprint arXiv:2501.01945}} (\bibinfo{year}{2025}).
\newblock


\bibitem[Zhang et~al\mbox{.}(2024)]%
        {PLoRA}
\bibfield{author}{\bibinfo{person}{You Zhang}, \bibinfo{person}{Jin Wang}, \bibinfo{person}{Liang-Chih Yu}, \bibinfo{person}{Dan Xu}, {and} \bibinfo{person}{Xuejie Zhang}.} \bibinfo{year}{2024}\natexlab{}.
\newblock \bibinfo{title}{Personalized LoRA for Human-Centered Text Understanding}.
\newblock
\newblock
\showeprint[arxiv]{2403.06208}~[cs.CL]
\urldef\tempurl%
\url{https://arxiv.org/abs/2403.06208}
\showURL{%
\tempurl}


\bibitem[Zhuang et~al\mbox{.}(2024)]%
        {zhuang2024hydra}
\bibfield{author}{\bibinfo{person}{Yuchen Zhuang}, \bibinfo{person}{Haotian Sun}, \bibinfo{person}{Yue Yu}, \bibinfo{person}{Rushi Qiang}, \bibinfo{person}{Qifan Wang}, \bibinfo{person}{Chao Zhang}, {and} \bibinfo{person}{Bo Dai}.} \bibinfo{year}{2024}\natexlab{}.
\newblock \showarticletitle{Hydra: Model factorization framework for black-box llm personalization}.
\newblock \bibinfo{journal}{\emph{Advances in Neural Information Processing Systems}}  \bibinfo{volume}{37} (\bibinfo{year}{2024}), \bibinfo{pages}{100783--100815}.
\newblock


\end{thebibliography}










\end{document}